\pdfoutput=1
\documentclass[11pt]{article}

\usepackage{multicol}

\usepackage{amsmath}
\usepackage{authblk}
\usepackage{multirow}
\usepackage{booktabs}
\usepackage{textcomp}
\usepackage{graphicx}
\usepackage{pdflscape}

\usepackage{emnlp2021}
\usepackage{times}
\usepackage{latexsym}

\usepackage[T1]{fontenc}
\usepackage[utf8]{inputenc}

\usepackage{microtype}

\title{GPT3Mix: Leveraging Large-scale Language Models \\ for Text Augmentation}

\author[1,2]{\textbf{Kang Min Yoo}}
\author[2]{\textbf{Dongju Park}}
\author[2]{\textbf{Jaewook Kang}}
\author[1,2]{\authorcr\textbf{Sang-Woo Lee}}
\author[2]{\textbf{Woomyeong Park}}
\affil[1]{NAVER AI Lab}
\affil[2]{NAVER Clova AI}
\affil[ ]{\texttt{\{kangmin.yoo, dongju.park, jaewook.kang\}@navercorp.com}}
\affil[ ]{\texttt{\{sang.woo.lee, max.park\}@navercorp.com}}

\begin{document}
\maketitle
\begin{abstract}

Large-scale language models such as GPT-3 are excellent few-shot learners, allowing them to be controlled via natural text prompts. Recent studies report that prompt-based direct classification eliminates the need for fine-tuning but lacks data and inference scalability. This paper proposes a novel data augmentation technique that leverages large-scale language models to generate realistic text samples from a mixture of real samples. We also propose utilizing soft-labels predicted by the language models, effectively distilling knowledge from the large-scale language models and creating textual perturbations simultaneously. We perform data augmentation experiments on diverse classification tasks and show that our method hugely outperforms existing text augmentation methods. We also conduct experiments on our newly proposed benchmark to show that the augmentation effect is not only attributed to memorization. Further ablation studies and a qualitative analysis provide more insights into our approach.

\end{abstract}

\section{Introduction}

% - Conventional NLP pipelines

% \input{Introduction/03_comment_Sangwoo}

In the seminal work by \citet{brown2020language}, a large-scale language model, specifically GPT-3, has been shown to achieve superior performance on zero-shot and few-shot learning tasks by prompt-based in-context learning.
In-context learning utilizes a prompt, which usually consists of a task description and few examples, to solve unseen tasks without the hefty price of fine-tuning. Recognizing the potential research applications of in-context learning and prompt-based control, a part of the NLP community has shifted its focus on understanding and devising advanced methods for optimizing prompt-based approaches \citep{schick2020exploiting, shin2020autoprompt, zhao2021calibrate, reynolds2021prompt}. 
% - Large-scale language models prohibits efficient training of downstream models
% - Using large-scale language models as the pre-trained language model has become prohitively expensive

However, these prompt-based approaches with inference on a large-scale language model suffer from several drawbacks.
First, the number of in-context training examples is hard limited by the maximum prompt length enabled by the inherent language model architecture.
Second, prompt-based approaches require online inference on the expensive large-scale language models. The inference may not be scalable in real-world use cases, because it is slow and incurs huge memory overhead.
Lastly, the prompt-based approaches do away with conventional machine learning techniques, making it mostly incompatible with existing established fine-tuning methods.

\begin{figure}[t]
    \centering
    \vspace{1.0cm}
    \includegraphics{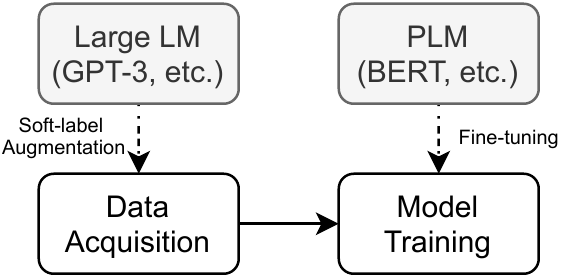}
    \caption{A conceptual diagram of text augmentation using large-scale language models.}
    \label{fig:framework}
\end{figure}

To overcome such limitations, we propose a more practical solution to utilize large-scale language models for downstream NLP tasks. In our proposed framework, as depicted in Figure \ref{fig:framework}, large-scale language models are not used as the pre-trained model for further domain-adaptive fine-tuning nor the backbone for prompt-based in-context learning but for imbuing the original training set with synthetic text data. 

We propose GPT3Mix, a method for generating synthetic but hyper-realistic text samples from a mixture of real samples utilizing large-scale language models such as GPT-3\footnote{Despite what the name suggests, we can apply GPT3Mix to any large-scale autoregressive language models.}. GPT3Mix extracts few sample sentences from the task-specific training data, embed these samples in the prompt, and generates an augmented mixed sentence influenced by the sample sentences. GPT3Mix uses soft-labels predicted by the large-scale language model to transfer knowledge of probability as in knowledge distillation \citep{hinton2015distilling}. In short, our method achieves both (1) \textit{data augmentation} via generating synthetic examples inspired by existing data samples and (2) \textit{knowledge distillation} by training smaller classification models using soft-labels predicted by the large language model. 

Our approach takes inspiration from the mix-based data augmentation methods in the vision domain \cite{zhang2017mixup}. Several mix-based data augmentation methods are suggested for NLP models. One of the notable methods is MixText \cite{chen2020mixtext}, in which BERT is used to generate novel augmentation samples from interpolated embedding spaces. However, despite its great success in the vision domain, deep-mixing text augmentation methods have seen limited effectiveness in real-world cases due to the difficulty of interpolating language from latent spaces \citep{bowman2016generating}. Synthetic language interpolated from a model's hidden space such as the word embedding space of BERT may introduce noise, outweighing the benefit of novel sample discovery and causing deterioration in the training data distribution. Our work exploits the generative power of large-scale language models like GPT-3 to generate high-quality mixed samples from in-context examples. 

We perform various data augmentation experiments on diverse classification tasks to verify our hypotheses and analyze our methodology. As language models are partly pretrained on web-crawled corpora, some benchmarks such as the movie review classification tasks may have been ``seen'' by the language models. To eliminate the possibility of pretraining memorization, we propose a new task \emph{RT20} where we collected online movie reviews posted after the known data preparation date of GPT-3. Experimental results with the newly proposed benchmark RT20 show that the benefit of our method is not attributed to memorization but mix-based text synthesis. We will release the benchmark soon.

The contribution of our work is summarized as follows.\footnote{The code to reproduce our results is available at https://github.com/naver-ai/hypermix.}

\begin{enumerate}
    \item We suggest employing prompt-based data augmentation using large-scale language models on top of the existing PLM fine-tuning paradigm to exploit the best of both worlds.
    \item We propose GPT3Mix, a simple but effective text augmentation technique, that elicits knowledge and linguistic capability possessed by large-scale language models.
    \item Our detailed analysis helps to understand the mechanism behind prompt-powered data augmentation, giving us insights into the generation and augmentation behavior.
    \item Our newly proposed RT20 task enables controlled experimentation on language models pretrained prior to a certain date, eliminating the possibility of memorization.
\end{enumerate}

\section{Related Work}

\paragraph{Knowledge Distillation} 

Knowledge distillation \citep{phuong2019towards} is a technique that trains a smaller student classifier on the outputs of a larger teacher classifier. Knowledge distillation for language models in the context of model compression has been well-studied in the literature. There have been various distilled models and distillation methods proposed for pre-trained language models \citep{sanh2019distilbert, tang2019distilling}. By utilizing soft-labels predicted by the large-scale language model, our approach helps to transfer knowledge to the downstream classifiers. 

\paragraph{Text Augmentation}

Text augmentation refers to methods for perturbing the linguistic space without altering class labels to improve the robustness and generalizability of the downstream models. Data augmentation has been studied extensively in the NLP scene. Text augmentation in the current literature comes with two flavors: shallow and deep augmentation. The shallow data augmentation techniques inject locally plausible small noises into the linguistic space (words or phrases), in the hopes that the perturbations produce linguistically acceptable samples while maintaining label consistency. Two examples are EDA \citep{wei2019eda} and synonym replacement \citep{zhang2016characterlevel}.

Another class of augmentation techniques employs external language models to improve global coherence and consistency. The back-translation approach exploits semantic consistency in translation language pairs to generate novel paraphrases \citep{fadaee2017data}. In the more recent line of work, pre-trained language models, such as BERT \citep{devlin2019bert} or the sequence-to-sequence variant BART \cite{lewis2020bart}, are used to obtain more diverse and linguistically correct augmentation samples. For example, BART has been proven to be effective in populating text samples for data-scarce labels \citep{kumar2020data}. \citet{ng2020ssmba} proposed using masked language models as a denoising autoencoder to generate synthetic texts. Some other researchers have taken the direction of perturbing the latent spaces, optionally by introducing variational inference in the architecture \citep{xia2020cg, xia2020composed, hou2018sequence, yoo2019data}.

On the other hand, inspired by the mix-up technique \citep{zhang2017mixup} proposed for the vision domain, there have also been works to mix existing text samples to produce realistic augmentation texts based on statistical methods \citep{guo2020sequence, sun2020mixup, chen2020mixtext}. Furthermore, pseudo-labeling, the act of annotating unlabeled data with model predictions \citep{lee2013pseudo, reed2014training}, has been actively used in semi-supervised learning settings \citep{chen2020mixtext, xie2020unsupervised, berthelot2019mixmatch}.

\paragraph{Large-scale Language Models}

Pre-trained transformer-based language models \citep{devlin2019bert, lewis2020bart} have initiated a new paradigm in the NLP scene, changing the way we design NLP pipelines. With the recent development of mega-scale language models \cite{shoeybi2019megatron, brown2020language}, we are witnessing another shift in the paradigm, namely prompt-based NLP. These large language models are essentially few-shot learners, allowing them to be controlled through natural text. There has been a steep rise in the community's interest to better understand the prompt-based mechanisms \citep{reynolds2021prompt, schick2020exploiting, shin2020autoprompt, jiang2020can, zhao2021calibrate}. Our work relies on the previous findings on prompt-based manipulation.

To the best of our knowledge, this work is the first to propose using the prompt-based approach to generate synthetic samples from large-scale language models for the purpose of text augmentation.

\section{GPT3Mix}
\label{sec:gpt3mix}

Mixup \citep{zhang2017mixup} is a simple learning technique that has been shown
to be effective in preventing memorization and improving generalizability 
for the vision domain.
The technique has been very effective on image data, 
but it has been harder to establish a standard approach 
for texts due to the inherent sparse nature of linguistic distributions,
which attributes to the challenges of identifying adversarial text examples
\citep{li2017robust}.
Inspired by the technique, we propose GPT3Mix as a powerful yet simple 
method to generate highly fluent synthetic samples based on a data
distribution. 

\begin{figure*}[t]
    \centering
    \includegraphics[scale=0.88]{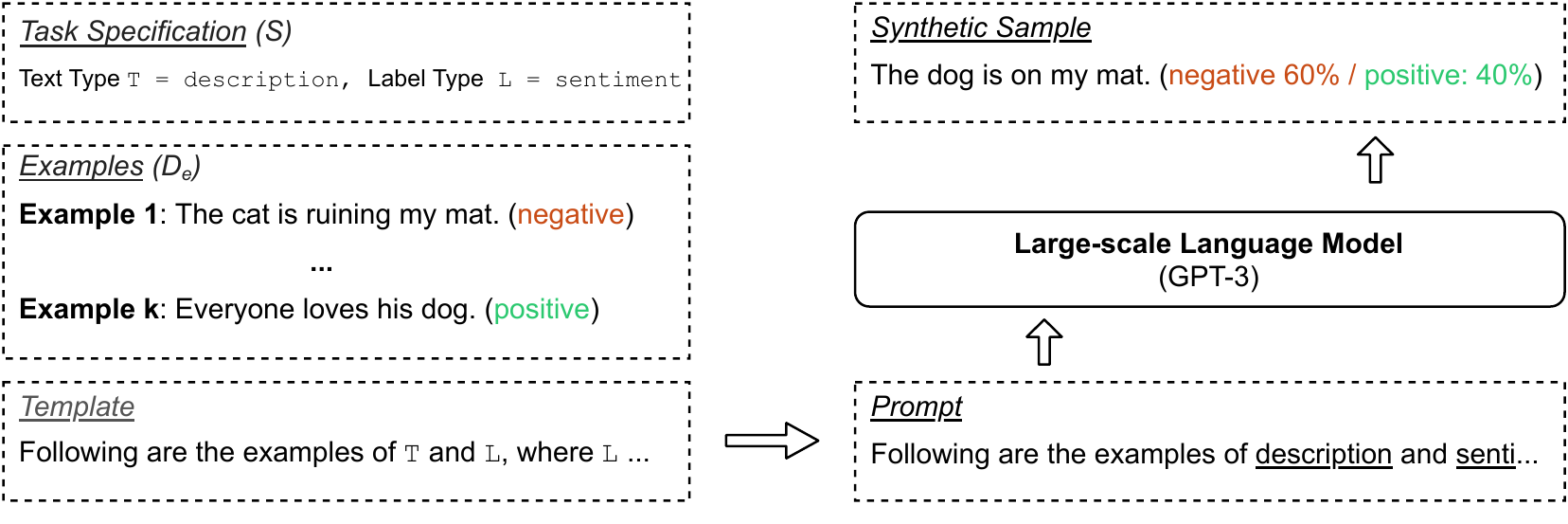}
    \caption{An illustration of GPT3Mix. The soft-labels of augmentation are extracted from the normalized label-token distributions predicted by the language model. Note that $v$ has been omitted in the task specification $S$ due to space limits.}
    \label{fig:gpt3mix}
\end{figure*}

The proposed method (Figure \ref{fig:gpt3mix}) consists of three steps: (1) selecting examples from the 
dataset, (2) constructing a GPT3Mix prompt from the selected examples and 
meta-information about the dataset, and finally (3) extracting
augmentation from the language model generation.
This section provides details about each step as follows.

\paragraph{Example Selection}

For simplicity, we confine the downstream task to text classification tasks. 
Given a classification task $\mathcal{T}$,
the training dataset $\mathcal{D}$ is a set of text $\mathbf{x}$ 
and associated label $y$ pairs: $\mathcal{D} = \left\{ \left( \mathbf{x}_i, y_i \right) \mid 1 \leq i \leq N \right\}$.

We randomly choose $k$ examples from $\mathcal{D}$ to be anchors.
Large-scale language models are known to be highly sensitive to 
the choice and the order of examples in the prompt \cite{reynolds2021prompt,zhao2021calibrate}.
We conjecture that by carefully choosing the examples, we are able to control
the generated augmentation samples from the language model.
We conduct qualitative analysis on the augmentation samples to 
confirm our hypothesis (\S \ref{sec:qualitative}).

In our implementation, we simply used uniform distribution to choose 
$k$ examples: $p_s (i) = 1 / N$.
Otherwise stated, most experiments are carried out by setting $k=2$ 
to simulate Mix-up.
As found in our ablation studies (\S \ref{sec:ablation-numex}), $k=2$
provides a good trade-off between cost and performance.

\paragraph{Prompt Construction}

Given a set of prompt examples $\mathcal{D}_e = \left\{ \left( \mathbf{x}_i, y_i \right) \mid 1 \leq i \leq k \right\}$ sampled from $\mathcal{D}$,
we formulate the prompt as follows.

A GPT3Mix prompt consists of a description header, an enumeration of text-label pairs of $\mathcal{D}_e$, and the augmentation prefix.
An example of the prompt is shown in the appendix (Appendix \ref{sec:app-prompts}).
Our prompt has been designed carefully with the current literature findings of
GPT-3 prompts \citep{reynolds2021prompt} in mind.

Specifically, the prompt follows the general template shown in the appendix,
but has task-specific information to allow the large-scale language models
to generalize better about the data distribution.
These task indicators are unique to each task and provide meta-information
of the task.
\begin{enumerate}
    \item \textbf{Text Type $T$}: Meta-type of the input text $\mathbf{x}$. For example, in movie review sentiment analysis, the text type corresponds to \verb|movie review|.
    \item \textbf{Label Type $L$}: Meta type of the label class $y$. For the example above, the label type corresponds to \verb|sentiment|.
    \item \textbf{Label-token Verbalizer $v: \mathcal{Y} \rightarrow \mathcal{V}$}: Similar to the concept of verbalizers in the work of \citet{schick2020s}, the 
    one-to-one mapping between the label classes $y \in \mathcal{Y}$ and word tokens in 
    the language model's vocabulary $\mathcal{V}$
    \footnote{In our implementation, we do not consider cases where a label class corresponds to multiple tokens. Regardless, expanding our work to incorporate multiple label tokens should be trivial.} is needed to formulate the prompt.
\end{enumerate}

The triple of the meta information above forms the task specification 
$S = \left( T, L, v \right)$.
Each task $\mathcal{T}$ requires a task specification $S_\mathcal{T}$ to be able to formulate a prompt for GPT3Mix. 
By default, the generic task specification 
$S_\text{generic} = \left( \text{text}, \text{label}, I \right)$
is used to construct prompts, where $I$ is the identity function assuming
that the class label exists in the vocabulary $\mathcal{V}$.

\paragraph{Augmentation Extraction}

The augmentation text $\mathbf{x}'$ and the label $y'$ are generated in succession after the prompt as a natural text. 
A predefined prompt template in the examples signals the language model to generate $\left( \mathbf{x}', y' \right)$ with a structure, allowing us to extract respective values through pattern matching.
Joint text and label generation also constraints the generated text to be associated with the correct label.

As illustrated in the prompt exhibit (Appendix \ref{sec:app-prompts}), our particular prompt design ensures that the label token that corresponds to $v\left( y' \right)$ is generated after $\mathbf{x}$.
This approach is inspired by the findings that, when inducing language models to come to a verdict, they require sufficient token lengths of ``silent reasoning'' prior to coming to a conclusion.

As large-scale language models are known to be few-shot learners \citep{brown2020language}, we also leverage GPT-3 to perform pseudo-labeling.
The likelihood of generating the label-tokens is normalized to obtain the soft-label probability of the augmentation text $\mathbf{x}'$.
Concretely, the pseudo-label probability of an augmentation
text $\mathbf{x}'$ being labelled with label $y'$ is as follows:

\begin{table*}[t]
    \centering
    \footnotesize
    \begin{tabular}{c|c|cccc|ccccc}
        \toprule
         & & \multicolumn{4}{c|}{\textbf{DistilBERT}\textsubscript{base}} & \multicolumn{5}{c}{\textbf{BERT}\textsubscript{base}} \\
         \midrule
         \textbf{Dataset} & \textbf{Sub.} & - & EDA & BT & Ours & - & EDA & BT & TMix & Ours \\
         \midrule \multirow{3}{*}{\textbf{SST-2}}
& $0.1\%$ & $56.6_{4.6}$ & $56.7_{6.8}$ & $56.9_{5.6}$ & $\mathbf{75.3}_{4.5}$ & $57.1_{4.6}$ & $56.6_{4.3}$ & $55.6_{3.8}$ & $56.9_{5.5}$ & $\mathbf{78.0}_{4.1}$ \\
& $0.3\%$ & $62.8_{6.2}$ & $63.1_{7.6}$ & $62.7_{5.8}$ & $\mathbf{82.1}_{2.2}$ & $65.6_{5.9}$ & $66.7_{5.2}$ & $66.5_{6.4}$ & $64.1_{7.6}$ & $\mathbf{84.9}_{1.4}$ \\
& $1.0\%$ & $79.2_{3.5}$ & $76.9_{2.3}$ & $77.4_{3.8}$ & $\mathbf{85.7}_{0.6}$ & $82.0_{2.8}$ & $79.6_{1.9}$ & $80.7_{3.1}$ & $79.9_{2.9}$ & $\mathbf{87.7}_{0.6}$ \\
\midrule \multirow{3}{*}{\textbf{COLA}}
& $0.1\%$ & $62.9_{6.3}$ & $57.3_{8.4}$ & $55.6_{6.0}$ & $\mathbf{68.6}_{0.1}$ & $60.7_{7.9}$ & $60.1_{6.8}$ & $55.2_{8.3}$ & $61.5_{8.9}$ & $\mathbf{68.6}_{0.2}$ \\
& $0.3\%$ & $64.1_{5.7}$ & $58.2_{4.4}$ & $54.7_{7.5}$ & $\mathbf{68.5}_{0.3}$ & $65.5_{5.0}$ & $63.0_{4.3}$ & $54.2_{6.5}$ & $67.9_{2.3}$ & $\mathbf{68.7}_{0.6}$ \\
& $1.0\%$ & $67.1_{2.3}$ & $59.8_{6.3}$ & $55.5_{5.9}$ & $\mathbf{68.6}_{0.3}$ & $\mathbf{70.9}_{2.3}$ & $63.2_{4.7}$ & $56.6_{6.4}$ & $70.2_{2.0}$ & $68.5_{0.3}$ \\
\midrule \multirow{3}{*}{\textbf{TREC6}}
& $0.1\%$ & $30.0_{7.2}$ & $30.4_{9.0}$ & $27.3_{6.7}$ & $\mathbf{41.3}_{5.3}$ & $32.1_{6.4}$ & $29.3_{7.1}$ & $30.3_{7.7}$ & $31.9_{8.2}$ & $\mathbf{47.7}_{7.5}$ \\
& $0.3\%$ & $39.3_{9.2}$ & $37.8_{8.0}$ & $40.4_{10.8}$ & $\mathbf{47.9}_{4.1}$ & $40.7_{9.2}$ & $42.0_{8.1}$ & $39.1_{11.5}$ & $39.3_{6.5}$ & $\mathbf{57.8}_{8.8}$ \\
& $1.0\%$ & $66.9_{5.8}$ & $62.6_{8.6}$ & $\mathbf{69.4}_{7.8}$ & $57.4_{2.8}$ & $67.0_{7.5}$ & $65.9_{7.1}$ & $69.3_{6.3}$ & $\mathbf{69.4}_{7.8}$ & $60.5_{6.1}$ \\
\midrule \multirow{3}{*}{\textbf{CR}}
& $0.1\%$ & $58.0_{4.7}$ & $58.9_{7.9}$ & $58.5_{7.9}$ & $\mathbf{69.2}_{6.3}$ & $59.0_{4.5}$ & $57.9_{7.1}$ & $57.9_{4.5}$ & $58.9_{5.6}$ & $\mathbf{70.0}_{5.8}$ \\
& $0.3\%$ & $63.1_{4.8}$ & $64.4_{5.2}$ & $61.4_{5.6}$ & $\mathbf{78.9}_{3.2}$ & $63.5_{6.6}$ & $65.3_{4.5}$ & $64.2_{5.5}$ & $63.0_{4.7}$ & $\mathbf{80.8}_{2.4}$ \\
& $1.0\%$ & $70.8_{5.7}$ & $71.7_{5.4}$ & $70.6_{4.6}$ & $\mathbf{83.2}_{1.2}$ & $75.8_{4.0}$ & $73.9_{3.5}$ & $74.6_{3.7}$ & $72.5_{4.6}$ & $\mathbf{84.7}_{1.9}$ \\
\midrule \multirow{3}{*}{\textbf{SUBJ}}
& $0.1\%$ & $\mathbf{83.9}_{2.5}$ & $83.8_{3.5}$ & $81.4_{5.2}$ & $82.3_{6.0}$ & $84.1_{4.0}$ & $84.7_{3.1}$ & $81.4_{7.2}$ & $83.6_{4.4}$ & $\mathbf{85.4}_{4.3}$ \\
& $0.3\%$ & $88.4_{1.0}$ & $\mathbf{88.4}_{1.3}$ & $87.2_{1.3}$ & $87.5_{1.5}$ & $89.3_{1.4}$ & $89.4_{3.5}$ & $88.4_{1.9}$ & $\mathbf{89.7}_{1.3}$ & $87.5_{2.3}$ \\
& $1.0\%$ & $\mathbf{90.7}_{0.9}$ & $90.5_{0.9}$ & $90.1_{0.7}$ & $89.3_{1.5}$ & $\mathbf{91.8}_{0.8}$ & $91.4_{1.1}$ & $90.9_{0.9}$ & $91.7_{0.9}$ & $90.6_{1.1}$ \\
\midrule \multirow{3}{*}{\textbf{MPQA}}
& $0.1\%$ & $66.5_{6.0}$ & $69.2_{5.0}$ & $62.3_{9.1}$ & $\mathbf{80.1}_{3.7}$ & $65.0_{4.7}$ & $69.1_{4.8}$ & $61.0_{6.8}$ & $65.2_{5.2}$ & $\mathbf{77.9}_{5.0}$ \\
& $0.3\%$ & $77.1_{5.4}$ & $78.2_{4.8}$ & $72.9_{6.8}$ & $\mathbf{85.0}_{0.9}$ & $71.3_{5.6}$ & $75.8_{3.5}$ & $72.6_{5.9}$ & $74.2_{3.4}$ & $\mathbf{84.7}_{1.0}$ \\
& $1.0\%$ & $84.0_{2.3}$ & $82.3_{2.9}$ & $82.2_{1.9}$ & $\mathbf{86.0}_{1.0}$ & $83.0_{3.4}$ & $81.9_{1.9}$ & $83.0_{2.4}$ & $83.0_{2.4}$ & $\mathbf{86.8}_{1.1}$ \\
\midrule \multirow{3}{*}{\textbf{RT20}}
& $0.1\%$ & $51.9_{2.6}$ & $52.1_{2.8}$ & $51.5_{2.6}$ & $\mathbf{55.0}_{5.3}$ & $50.9_{2.1}$ & $51.8_{2.7}$ & $53.1_{2.6}$ & $53.2_{3.7}$ & $\mathbf{57.1}_{4.1}$ \\
& $0.3\%$ & $51.9_{2.5}$ & $51.6_{3.0}$ & $51.2_{2.2}$ & $\mathbf{60.7}_{4.5}$ & $51.4_{2.7}$ & $51.9_{2.8}$ & $51.4_{3.6}$ & $52.0_{2.6}$ & $\mathbf{65.0}_{5.2}$ \\
& $1.0\%$ & $56.2_{5.7}$ & $55.0_{3.6}$ & $55.9_{4.1}$ & $\mathbf{72.3}_{1.9}$ & $57.9_{4.5}$ & $57.9_{5.3}$ & $57.4_{4.2}$ & $56.0_{4.4}$ & $\mathbf{75.4}_{2.5}$ \\
    \midrule \multirow{3}{*}{\textbf{Average}}
& $0.1\%$ & $58.5$ & $58.3$ & $56.2$ & $\mathbf{67.4}$ & $58.4$ & $58.5$ & $56.4$ & $58.7$ & $\mathbf{69.2}$ \\
& $0.3\%$ & $63.8$ & $63.1$ & $61.5$ & $\mathbf{72.9}$ & $63.9$ & $64.9$ & $62.4$ & $64.3$ & $\mathbf{75.6}$ \\
& $1.0\%$ & $73.6$ & $71.2$ & $71.6$ & $\mathbf{77.5}$ & $75.5$ & $73.4$ & $73.2$ & $74.7$ & $\mathbf{79.2}$ \\
    \bottomrule 
    \end{tabular}
    \caption{Main data augmentation results on 0.1\%, 0.3\%, and 1.0\% training set sub-sample levels. We compare different augmentation strategies by transformer architectures on the downstream classification performance. Experiments have been repeated 10 times and the statistics are presented in the mean\textsubscript{std} format.}
    \label{tab:exp-main}
\end{table*}

\begin{equation}
    p \left( y' \mid \mathbf{x}' \right) \propto p_{\text{LM}} \left( v_\mathcal{T} \left( y' \right) \mid \mathcal{P} \left( \mathbf{x}', S_\mathcal{T} \right) \right) ,
\end{equation}

\noindent where $p_\text{LM}$ is the language modeling likelihood and $\mathcal{P} : \mathcal{S} \rightarrow \mathcal{X}$ is the function that constructs the prompt given a task specification.

Our approach effectively combines text perturbation, pseudo-labeling, and knowledge distillation in a single augmentation operation. 
In practice, augmentation samples with pseudo-labels are trained along with the real samples using the cross-entropy loss.
This is in contrast to prior work, in which pseudo-labels are usually used for consistency regularization in the context of semi-supervised learning \citep{berthelot2019mixmatch}. 

\section{Experiments}

We evaluate our augmentation approach on the following seven classification benchmarks: 

\textbf{SST-2} \citep{socher2013recursive} is a sentiment classification dataset that contains movie reviews crawled from Rotten Tomatoes and their corresponding binary labels. \textbf{CR} \citep{hu2004mining} dataset is a set of Amazon product reviews labeled by binary sentiments. The Corpus of Linguistic Acceptability (\textbf{COLA}) \citep{warstadt2018neural} is a collection of sentences extracted from publications annotated with grammaticality. The \textbf{TREC6} dataset \citep{voorhees1999trec} concerns the question classification task consisting of open-domain, fact-based questions divided into broad semantic categories. \textbf{MPQA} \citep{wiebe2005annotating} consists of opinions and their semantic polarity. The subjectivity dataset (\textbf{SUBJ}) \citep{pang2004sentimental} contains movie reviews labeled with objectivity. 

\textbf{RT20} is the newly proposed benchmark with which we perform controlled experiments on language models. The dataset is a binary sentiment classification corpus, collected from Rotten Tomatoes accessed after a certain date. The details about the collection and preparation is provided in Appendix \ref{sec:app-rt20}.

% The statistics of the datasets are summarized in table \ref{tab:exp-data}.

\subsection{Experimental Settings}

To showcase our approach, we conduct downstream classification experiments on artificially data-scarce tasks by sub-sampling the training set.
For each experiment, we perform a class-balanced sub-sample on the training set.
We account for statistical variance in our experiments by fixating the sub-samples on 15 different data seeds and repeating the augmentation procedure and downstream classification experiments on all sub-samples.
The data seeds were chosen randomly\footnote{The data seeds were randomly generated using a master seed.}.

For the classifier architecture, we use the base size BERT \citep{devlin2019bert} and DistilBERT \citep{sanh2019distilbert} models, which have 109M and 67M parameters respectively. For each downstream classification trial, we initialize the classifier model with the pre-trained parameters provided by the Huggingface Transformers library \citep{wolf2019huggingface} and randomly initialize the classifier layers, which consist of two fully connected layers that predict the class labels from the output embeddings of the transformer architectures. The classifiers are trained automatically by employing early stopping against the validation score with patience of 20 training epochs. We report classification accuracies in all of our tables.

\subsection{Implementation Details}

For selecting the optimal task specification for each task in GPT3Mix augmentation, we evaluated the performance of few handcrafted task specification candidates on the validation set and chose the highest performing one. The details about the optimal task specifications are presented in Appendix \ref{sec:app-meta}. The inference on GPT-3 was carried out via the OpenAI API Beta Access program. We used the largest GPT-3 model available on (\verb|davinci|) unless otherwise stated. On average, a GPT3Mix augmentation roughly consumes 300 tokens in combined length (prompt and generation). For GPT-3 generation, top-p and the temperature was set to 1 and the frequency penalty was set to 0.02 \citep{holtzman2019curious}. The augmentation ratio between the training set and the synthetic set was set to 10 unless otherwise stated.

During classifier training, we used the Adam optimizer with decoupled weight decay \citep{kingma2014adam, loshchilov2017decoupled} and a learning rate of 3e-5. The learning rate had a warm-up period of 3 epochs. PyTorch and M40 GPUs were used to run the experiments.

\subsection{Data Augmentation Experiments}
\label{sec:exp-da}

\begin{table}[t]
    \centering
    \footnotesize
    \begin{tabular}{c|cccc}
        \toprule
         & \multicolumn{4}{c}{\textbf{BERT}\textsubscript{large}} \\
         \midrule
         \textbf{Sub.} & - & EDA & BT & Ours \\
         \midrule
         $0.1\%$ & $60.3_{7.9}$ & $63.8_{7.2}$ & $63.4_{7.4}$ & $\mathbf{84.0}_{4.4}$ \\
         $0.3\%$ & $74.1_{8.9}$ & $73.2_{5.8}$ & $73.1_{9.4}$ & $\mathbf{88.7}_{1.0}$ \\
         $1.0\%$ & $87.8_{1.5}$ & $87.3_{1.4}$ & $87.0_{2.7}$ & $\mathbf{90.8}_{0.6}$ \\
        \bottomrule 
    \end{tabular}
    \caption{Additional data augmentation experiments on SST-2 with BERT\textsubscript{large}, which has 335M parameters. The larger network capacity enables the model to better exploit the GPT3Mix augmentations, allowing it to match the performance of BERT\textsubscript{base} trained on the full data with just 1.0\% subsample of the training data.}
    \label{tab:exp-main-large}
\end{table}

We compare our approach to Easy Data Augmentation (EDA) \cite{wei2019eda}, back-translation (BT) \citep{fadaee2017data}, and TMix \cite{chen2020mixtext}. For the back-translation baseline, texts were translated to and from German using Transformer architectures trained on the WMT16 English-German corpus provided by Fairseq \citep{ott2019fairseq}. For TMix, we employ the hyperparameters reported by the authors. We compare with TMix on BERT\textsubscript{base} only, since the effectiveness of TMix is not established in other architectures\footnote{Our attempt on searching TMix hyperparameters for DistilBERT\textsubscript{base} and BERT\textsubscript{large} did not yield meaningful results.}.

\begin{table}[t]
    \centering
    \footnotesize
    \begin{tabular}{c|cccc}
        \toprule
         & \multicolumn{4}{c}{$k$} \\
         \midrule
         \textbf{Sub.} & 1 & 2 & 4 & 8  \\
         \midrule
         $0.1\%$ & 
         $65.5_{3.3}$ & $71.2_{6.5}$ & $74.6_{3.9}$ & $72.0_{6.7}$ \\
         $0.3\%$ & 
         $78.9_{3.9}$ & $80.0_{2.7}$ & $80.2_{2.1}$ & $80.0_{1.6}$ \\
         $1.0\%$ &
         $85.2_{0.6}$ & $84.3_{0.7}$ & $84.3_{0.7}$ & $84.2_{1.2}$ \\
         \bottomrule
    \end{tabular}
    \caption{An ablation study on the number of examples $k$ in GPT3Mix prompts. When $k=1$, GPT-3 produces point-wise perturbed samples. Experiments are carried out on the SST-2 dataset.}
    \label{tab:ablation-numex}
\end{table}

The results on data-scarce text augmentation are presented in Table \ref{tab:exp-main}. First, we notice that, in most cases, our approach outperforms other augmentation baselines by a large margin. Also, our approach achieves higher stability in terms of the variance of repeated trials and inter-task fluctuations than other augmentation methods. Although back-translation and EDA do outperform GPT3Mix in certain configurations, GPT3Mix offers the most consistent performance boost for the downstream classifiers across all tasks. This is evident from the average classification accuracies of all tasks, in which GPT3Mix improves the baseline as much as 18.6\% (for BERT\textsubscript{base}) while other methods show nearly no improvement\footnote{We employed the hyperparameters proposed by the authors of EDA and BT.}.

We also note that, despite non-augmented baselines of DistilBERT\textsubscript{base} and BERT\textsubscript{base} being very close (58.5 and 58.4 respectively on average of 0.10\% subsamples), a much larger augmentation effect is observed in BERT\textsubscript{base} results (67.4 \textrightarrow 69.2). Improving model robustness is known to require significantly larger model complexity \cite{ye2019adversarial}, hence BERT\textsubscript{base}, having 65\% more parameters than DistilBERT\textsubscript{base}, utilizes GPT3Mix samples better than the counterpart. This effect is more apparent in the even larger model (Table \ref{tab:exp-main-large}), which outperforms fully trained BERT\textsubscript{base} with just 1\% of the original data.

Furthermore, we observe that augmenting with GPT3Mix significantly improves the baseline across all subsamples of RT20, eliminating the suspicion that the data augmentation effect of GPT3Mix is attributed to data memorization of GPT-3. Also note that, due to the recency of the RT20 dataset, the pretrained classification transformers do not perform as well as on the older counterpart, SST-2. However, GPT3Mix is able to alleviate the difficulty through knowledge distillation and mix-based robust training.

\paragraph{Full Dataset Experiments}

We also perform full dataset data augmentation experiments to confirm that GPT3Mix still offers benefits even when task-specific data are abundant. We augmented the full SST-2 dataset with one-to-one ratio of synthetic samples from GPT3Mix, and the experiments show that GPT3Mix improves the accuracy of DistilBERT\textsubscript{base} from $90.28\%$ to $90.70\%$ ($0.42\%$ improvement) and the accuracy of BERT\textsubscript{base} from $90.33\%$ to $93.25\%$ ($2.92\%$ improvement). Again, we observe a larger improvement in the more expressive model, in align with previous findings \citep{zhang2017mixup, shafahi2019adversarial}.

\subsection{Ablation Studies}
\label{sec:ablation}

We conduct a number of ablation experiments to study the underlying mechanism of GPT3Mix. Note that the augmentation results for GPT3mix in the following ablation studies may underperform compared to the results presented in \S \ref{sec:exp-da} due to ablation studies having lower augmentation ratios and using smaller language models (\verb|curie|). Also note that all ablation experiments were carried out on the DistilBERT\textsubscript{base} classifier architecture.

\subsubsection{Number of Prompt Examples}
\label{sec:ablation-numex}

\begin{table}[t]
    \centering
    \footnotesize
    \begin{tabular}{c|cccc}
        \toprule
        & \multicolumn{4}{c}{\textbf{Model Size}} \\
        \midrule
         \textbf{Sub.} & {\small \verb|ada|} & {\small \verb|babbage|} &  {\small \verb|curie|} & {\small \verb|davinci|}  \\
         \midrule
         $0.1\%$ & 
         $61.9_{4.1}$ & $65.2_{6.9}$ & $65.9_{5.3}$ & $67.6_{7.2}$ \\
         $0.3\%$ & 
         $74.6_{4.8}$ & $69.7_{7.3}$ & $74.6_{4.5}$ & $78.3_{2.9}$ \\
         $1.0\%$ &
         $81.6_{1.0}$ & $82.5_{1.1}$ & $83.4_{1.8}$ & $84.3_{1.1}$ \\
         \bottomrule
    \end{tabular}
    \caption{An ablation study on the size of the language model with the SST-2 dataset. Larger language models provide greater augmentation benefits in data-limited environment.}
    \label{tab:ablation-engine}
\end{table}

\begin{table}[t]
    \centering
    \footnotesize
    \begin{tabular}{c|cccc}
        \toprule
        & \multicolumn{4}{c}{\textbf{Pre-trained Language Model}} \\
        \midrule
         \textbf{Sub.} & - & {\small GPT-2} & {\small GPT-neo} & {\small \verb|davinci|}  \\
         \midrule
         $0.1\%$ & $56.6_{4.6}$ &
         $64.1_{6.5}$ & $71.3_{4.7}$ & $\mathbf{75.3}_{4.5}$ \\
         $0.3\%$ & $62.8_{6.2}$ &
         $76.9_{3.6}$ & $80.2_{1.9}$ & $\mathbf{82.1}_{2.2}$ \\
         $1.0\%$ & $79.2_{3.5}$ &
         $76.1_{3.6}$ & $82.6_{1.1}$ & $\mathbf{85.7}_{0.6}$ \\
         \bottomrule
    \end{tabular}
    \caption{ Open-source alternatives are compared to the largest GPT-3 model on the SST-2 dataset. For GPT-2, the large version that has 774M parameters was used. For GPT-neo, the smaller version of 1.3B parameters was used.}
    \label{tab:ablation-open-source}
\end{table}

\begin{table}[t]
    \centering
    \footnotesize
    \setlength\tabcolsep{4.5pt}
    \begin{tabular}{c|l}
        \toprule
        \textbf{Example 1} & Laughably, irredeemably awful. (\textit{negative}) \\
        \textbf{Example 2} & Well-made but mush-hearted. (\textit{positive}) \\
        \midrule
        \textbf{GPT3Mix} & Groundbreaking, disturbing. \\ & (\textbf{\textit{positive}}: 75\%, \textit{negative}: 25\%) \\
        \midrule \midrule
        \textbf{Example 1} & It's just not very smart. (\textit{negative}) \\
        \textbf{Example 2} & It's quite an achievement to set and shoot \\
        & a movie at the Cannes Film Festival and yet \\
        & fail to capture its visual appeal or \\
        & its atmosphere. (\textit{negative}) \\
        \midrule
        \textbf{GPT3Mix} & Excessively talky, occasionally absurd and \\
        & much too long, Munich is a fascinating \\
        & mess. \\
        & (\textit{positive}: 21\%, \textbf{\textit{negative}}: 79\%) \\
        \bottomrule
    \end{tabular}
    \caption{SST-2 augmentation samples from GPT3Mix (davinci). GPT3Mix annotates synthetic samples with soft-labels predicted by the language model.}
    \label{tab:gpt3mix-samples}
\end{table}

First, the effect of the number of examples in GPT3Mix prompts ($k$) on the downstream augmentation performance is studied. GPT3Mix requires $k \geq 2$ to effectively mix existing samples and generate interpolated text samples. However, supplying one example ($k=1$) per prompt and expecting GPT-3 to introduce perturbations or paraphrases of the given example can be a viable strategy. We vary $k$ on the SST-2 dataset and observe the downstream performances (Table \ref{tab:ablation-numex}). The second-largest GPT-3 model (\verb|curie|) was used and the augmentation multiplier was set to 10.

From the results, we notice that when the data availability is severely limited (i.e. 0.1\% and 0.3\%), point-wise perturbation doesn't offer the performance improvement as much as when $k \geq 2$. However, as data becomes more abundant, increasing the number of mixing samples offers marginally small benefits for data augmentation. Yet, increasing the number of examples incurs additional overhead to the GPT-3 inference cost. 

Generally, over-providing prompt examples may constraint the degrees of freedom and causing the synthetic samples to overfit on the data, hurting the downstream performances. However, a significant improvement from $k=2$ to $k=4$ is observed for the $0.1\%$ sub-sample level. In our data augmentation studies, we weigh in on $k=2$ as a reasonable balance between the trade-off between GPT-3 inference costs and performance gains.

\subsubsection{Language Model Capacity}

\begin{table}[t]
    \centering
    \small
    \begin{tabular}{c|ccc}
        \toprule
         \textbf{Sub.} & No Aug. & Hard Labels & Soft-labels \\
         \midrule
         $0.1\%$ & $55.8_{5.1}$ & $61.6_{8.0}$ & $71.2_{6.5}$ \\
         $0.3\%$ & $64.9_{8.0}$ & $67.7_{5.9}$ & $80.0_{2.7}$ \\
         $1.0\%$ & $77.9_{3.6}$ & $79.0_{2.8}$ & $84.3_{0.7}$ \\
         \bottomrule
    \end{tabular}
    \caption{An ablation study on the employment of pseudo-labels. Hard labels are obtained from the beam search of the entire sequence autoregressively generated by the language model.}
    \label{tab:ablation-hard}
\end{table}

Next, we study the influence of the model capacity of the augmenting language model on the quality of augmentations. OpenAI offers GPT-3 in four different capacities: \verb|ada|, \verb|babbage|, \verb|curie|, and \verb|davinci|\footnote{The sizes of the language models are known to be 2.7B, 6.7B, 13B, and 175B respectively; however, OpenAI has not officially disclosed the exact numbers yet.}, listed in the increasing order of model complexity. In this study, the augmentation ratio is set to 5. The results (Table \ref{tab:ablation-engine}) show that having larger and more expressive language models benefit data augmentation.

{

Additionally, we conduct comparative experiments to verify whether open-source alternatives to GPT-3 could still provide comparable performance gains through data augmentation. As open-source alternatives, GPT-2 \cite{radford2019language} and GPT-neo \cite{gpt-neo} were chosen. The latter is a popular alternative to the commercial GPT-3, performing competitively with the smaller versions (\verb|ada| and \verb|babbage|) of the counterpart. Our results (Table \ref{tab:ablation-open-source}) show that the open-source GPT-like models still provide comparable performance gains, strongly suggesting that our prompt-based GPT3Mix approach can be versatile in the choice of pre-trained language models. Even the smaller GPT-2 model could provide performance gains.

}

\subsubsection{Task Specification}

We are also interested in how the design choice of task specification for prompt construction affects the downstream performance. To analyze the effect, we compare the optimal task specification $S_\mathcal{T^\star}$ to a generic one ($S_\text{generic}$), where the nature of the task cannot be inferred from the description.  For this study, we used \verb|curie| as the augmenting language model with an augmentation ratio of 3. The results in Table \ref{tab:ablation-task} support our conjecture that the language model utilizes the meta-information about the dataset to generate better data samples, and thus prompt designs have a significant impact on the augmentation quality. However, the generic task specification outperforms other augmentation baselines, highlighting the effectiveness of employing large-scale language models as the augmentation source.

\subsubsection{Pseudo-labeling}

\begin{table}[t]
    \centering
    \small
    \begin{tabular}{c|c|ccc}
        \toprule
         \textbf{Dataset} & \textbf{Sub.} & No Aug. & $S_\text{generic}$ & $S_\mathcal{T^\star}$ \\
         \midrule
         \multirow{3}{*}{\textbf{SST-2}} 
         & $0.1\%$ & $55.8_{5.1}$ & $60.1_{5.2}$ & $71.2_{6.5}$ \\
         & $0.3\%$ & $64.9_{8.0}$ & $72.6_{5.7}$ & $80.0_{2.7}$ \\
         & $1.0\%$ & $77.9_{3.6}$ & $81.4_{1.7}$ & $84.3_{0.7}$ \\
         \midrule
         \multirow{3}{*}{\textbf{COLA}} 
         & $0.1\%$ & $64.9_{4.7}$ & $68.4_{0.4}$ & $68.6_{0.0}$ \\
         & $0.3\%$ & $62.2_{7.2}$ & $65.7_{2.7}$ & $68.7_{0.2}$ \\
         & $1.0\%$ & $67.8_{1.6}$ & $68.7_{0.3}$ & $69.1_{1.1}$ \\
         \bottomrule
    \end{tabular}
    \caption{An ablation study on task specifications. $S_\text{generic}$ denotes a generic task specification that does not hold task-specific meta-information (\S \ref{sec:gpt3mix}), and $S_\mathcal{T^\star}$ denotes the optimal specification for the corresponding task.}
    \label{tab:ablation-task}
\end{table}

Finally, we study the effect of employing pseudo-labels from the label token probabilities predicted by the large-scale language model. we compare the augmentation performance when the label tokens optimized from the sequence-wide beam search are used instead. Results on SST-2 (Table \ref{tab:ablation-hard}) show that employing soft-labels has a strong advantage over sequence-optimized labels. The performance gap between the hard and soft-labels can be considered as the benefit of utilizing the class distribution jointly predicted by the language model as a form of knowledge distillation for synthetic samples \citep{kim2016sequence}. \verb|curie| was used as the GPT-3 model with the augmentation ratio of 5. 

\subsubsection{Qualitative Analysis}
\label{sec:qualitative}

Language models are known to be sensitive to the selection and the order of the examples presented in the prompt, causing biases in the predictions \citep{zhao2021calibrate, reynolds2021prompt}. Our proposed method hinges on this unique property of large-scale language models, hence we wish to qualitatively examine the augmentation samples to further support our hypothesis. 

The augmentation samples for the SST-2 dataset are presented in Table \ref{tab:gpt3mix-samples}. First, we notice that the synthetic sentiment is correlated with the input sentiments. If both examples are either all \textit{negative} (the second example), the sentiment of the augmentation sample is heavily biased towards \textit{negative}. Second, we also discover that the augmented sample follows the similar syntactic and semantic structure of the example texts. As demonstrated in the first case, the short and phrasal structure of the examples is well translated into the generated sample, supporting the notion that language models are able to learn from in-context examples even for generation and pseudo-labeling tasks. In the second example, the linguistic similarity between the generated sample and the given examples is more abstract (use of adjective phrases and enumerated clauses), suggesting that language models are capable of creative interpolation.

\section{Conclusion}

In this paper, we proposed a novel text augmentation technique that leverages large-scale language models and their abilities to perform controlled generation via prompts. Our extensive experiments on classification tasks show that our augmentation method can improve robustness of pretrained transformers through mix-based perturbation and knowledge distillation without the online inference on heavy LMs. Thus, our method can be a competitive alternative to prompt-based task-solving \citep{brown2020language} or direct fine-tuning \citep{liu2021gpt}. As future work, we are interested in the possibility of further pushing the boundaries of state-of-the-art architectures via GPT3Mix. We are also working towards improving generation efficiency by optimizing example selection and prompt templates.

{

\section{Ethical Considerations}

Our approach presents several ethical challenges. Pre-trained language models that are trained on untreated corpora are known to exhibit social biases \citep{bordia2019identifying, hutchinson2020social, abid2021persistent, bender2021dangers} and toxicity \citep{gehman2020realtoxicityprompts}. The biased property is concerning because language models are prone to degeneration even in the absence of bias or toxicity in the prompts \citep{gehman2020realtoxicityprompts}. As a result, GPT3Mix is not exempt from the possibility of propagating linguistic biases and toxicity even if the real training examples were ensured to be unbiased. Furthermore, linguistic bias could be amplified through iterative applications of GPT3Mix (i.e., using GPT3Mix-augmented samples as the source examples for the next iteration of GPT3Mix). 

To address these issues, we propose three remedies to reduce the concerns. First, debiased pre-trained language models can be used in place of GPT-3. Language models can be adapted to debiased and non-toxic corpora \cite{gehman2020realtoxicityprompts} or treated with modifications to the word embedding space \cite{basta2021impact} to inhibit their tendency to generate bias. Moreover, GPT3Mix has been shown to work well with various pre-trained language models (Table \ref{tab:ablation-open-source}). Second, specific decoding strategies can be employed to reduce bias at inference time. Recent body of work has shown that handcrafted dictionaries can be employed to suppress the selection of offensive words \cite{gehman2020realtoxicityprompts} and that language models can implicitly learn to identify biases through self-diagnosis, which can be exploited for self-debiasing \cite{schick2021self}. Third, human-in-the-loop in the augmentation process can be utilized to manually identify and filter linguistic bias.

Note that the ethical implications can be minimized by using GPT3Mix only for augmenting discriminators, where the augmented samples are removed once the training process is complete. However, for the general purpose of populating datasets, linguistic bias is of ethical concern and can be alleviated using the existing work on debiasing.
}
% Problem
% - 1: PLMs trained on openly crawled corpora may exhibit social biases (hutchinson2020social, abid2021persistent) and toxicity (gehman2020realtoxicityprompts)
%   - The work on Toxicity  shows that language models are prone to degeneration even when prompted with clean texts.
%   - This is a major concern to us, since gpt3mix may introduce ethically questionable samples even in the absence of bias in the real examples.
%   - Furthermore, biases exhibited by the PLMs could be amplified through iterative applications of GPT3Mix (i.e., using the GPT3Mix-augmented samples as the source examples for the next iteration of GPT3Mix).
  
% Solution
%  - We propose (1) using detoxified PLM instead (2) Using special decoding strategies (3) human-filtering
%  - However, if GPT3Mix is only used for augmenting discriminators, then the issue is not that apparent, we suggest using the augmented examples temporarily and deleting them after 
%  - However, we agree that when GPT3Mix is solely used for the purpose of populating datasets, linguistic biases are a concern and can be alleviated using the aforemented approaches

\section*{Acknowledgement}

We would like to show appreciation for the valuable feedback given by Jung-Woo Ha, Gyuwan Kim, and Hwaran Lee through detailed reviews. We also thank Jaimeen Ahn for providing comments on ethical considerations of our approach and the underlying language model. 


\bibliography{emnlp2021}
\bibliographystyle{acl_natbib}

\onecolumn
\appendix

\section{Prompts}
\label{sec:app-prompts}

The GPT3Mix prompt uses the following template. The template corresponds to the prompt-constructing function $\mathcal{P}$, which require a task specification $S_\mathcal{T} = ( T, L, v )$.

\begin{verbatim}
Each item in the following list contains a <text type> and the
    respective <label type>. <label type> is one of '<label token 1>',
    ..., or '<label token N>'.

<text type>: <example text 1> (<label type>: <example label 1>)
...
<text type>: <example text k> (<label type>: <example label k>)
<text type>:    
\end{verbatim}

For example, given $S_\text{SST2} = ( \text{movie~review}, \text{sentiment}, I )$, the constructed GPT3Mix prompt is as follows.

\begin{verbatim}
Each item in the following list contains a movie review and the
respective sentiment. The sentiment is one of 'positive' or 'negative'.

Movie review: Despite its Hawaiian setting, the science-fiction
    trimmings and some moments of rowdy slapstick, the basic plot of
    ``Lilo'' could have been pulled from a tear-stained vintage Shirley
    Temple script. (Sentiment: Negative)
Movie review: And people make fun of me for liking Showgirls.
    (Sentiment: Negative)
Movie review:
\end{verbatim}

\section{Task Specifications}
\label{sec:app-meta}

\begin{table}[h]
    \centering
    \small
    \begin{tabular}{c|ccc}
        \toprule
        \textbf{Dataset} & $T$ & $L$ & $v$ \\
        \midrule
        Generic & text & label & $\cdot \rightarrow \cdot$ \\
        SST-2 & movie review & sentiment & \verb|pos| \textrightarrow \verb|positive|, \verb|neg| \textrightarrow \verb|negative| \\
        CR & customer review & sentiment & \verb|pos| \textrightarrow \verb|positive|, \verb|neg| \textrightarrow \verb|negative| \\
        SUBJ & text & objective & \verb|subjective| \textrightarrow \verb|no|, \verb|objective| \textrightarrow \verb|yes| \\
        COLA & text & grammar & \verb|acceptable| \textrightarrow \verb|correct|, \verb|unacceptable| \textrightarrow \verb|incorrect| \\
        TREC6 & question & type & \verb|ABBR| \textrightarrow \verb|abbreviation|, \verb|LOC| \textrightarrow \verb|location|, \\
        & & & \verb|DESC| \textrightarrow \verb|description|, \verb|NUM| \textrightarrow \verb|numeric| \\
        & & & \verb|ENTY| \textrightarrow \verb|entity|, \verb|HUM| \textrightarrow \verb|human| \\
        MPQA & text & sentiment & \verb|pos| \textrightarrow \verb|positive|, \verb|neg| \textrightarrow \verb|negative| \\
        \bottomrule
    \end{tabular}
    \caption{Optimal task specifications.}
    \label{tab:app-task}
\end{table}

After validating candidate task specifications for each task, we have selected the following for conducting our experiments (Table \ref{tab:app-task}).

Providing incorrect or suboptimal specifications to the prompt may cause a large drop in augmentation qualities. For example, in the case of designing task specifications for the COLA dataset, when ``linguistic acceptability'' is used as the label type (instead of the optimal ``grammar''), the downstream performance on the 0.1\% sub-dataset drops to 38.8\%, resulting in performance worse than the non-augmented baseline of 68.80\%.

\section{RT20 Dataset}
\label{sec:app-rt20}
RT20 is a new binary sentiment classification dataset made up of movie reviews posted for movies released in 2020 or thereafter. This newly created dataset is free from the training dataset used by GPT-3, eliminating the possibility of performance improvement due to memorization. 

To build this dataset, we crawled critic reviews of movies released in or after 2020 that were included in the movie category on Rotten Tomatoes\footnote{https://www.rottentomatoes.com/}. Generally, the critic reviews have higher linguistic acceptability than user reviews, allowing us to control data quality. For each movie, fresh and rotten reviews were sampled at a 1:1 ratio, with ``positive'' being labeled for fresh reviews and ``negative'' being labeled for rotten reviews. During preprocessing, all characters were replaced with the lowercase letters, and spaces were added before and after certain special characters: ``".?!:()[],''. The final corpus is a collection of 1,100 positive and 1,100 negative reviews for 62 recent movies. We further split the dataset into 1500 training, 300 validation, and 400 test data using the class-balanced sampling strategy.

% Description of the dataset (Collection method etc.)

\section{GPT3Mix Samples}

\noindent The following GPT3Mix examples are generated using the largest GPT-3 model (\verb|davinci|) on SST-2.

\vspace{1em}

\begin{tabular}{c|l}
    \toprule
    \textbf{Example 1} & Laughably, irredeemably awful. (\textit{negative}) \\
    \textbf{Example 2} & Well-made but mush-hearted. (\textit{positive}) \\
    \midrule
    \textbf{GPT3Mix} & Groundbreaking, disturbing. (\textbf{\textit{positive}}: 75\%, \textit{negative}: 25\%) \\
    \midrule \midrule
    \textbf{Example 1} & Berry's saucy, full-bodied performance give this aging series a much needed kick, \\
    & making ``Die Another Day'' one of the most entertaining Bonds in years. (\textit{positive}) \\
    \textbf{Example 2} & Moonlight Mile doesn't quite go the distance but the cast is impressive and they all \\ 
    & give life to these broken characters who are trying to make their way through \\
    & this tragedy. (\textit{positive}) \\
    \midrule
    \textbf{GPT3Mix} & ``Gosford Park'' takes a leisurely approach to story telling and along the way it mixes \\
    & intrigue, comedy, and romance. (\textbf{\textit{positive}}: 98\%, \textit{negative}: 2\%) \\
    \midrule \midrule
    \textbf{Example 1} & It's just not very smart. (\textit{negative}) \\
    \textbf{Example 2} & It's quite an achievement to set and shoot a movie at the Cannes Film Festival \\
    & and yet fail to capture its visual appeal or its atmosphere. (\textit{negative}) \\
    \midrule
    \textbf{GPT3Mix} & Excessively talky, occasionally absurd and much too long, \\
    & Munich is a fascinating mess. (\textit{positive}: 21\%, \textbf{\textit{negative}}: 79\%) \\
    \bottomrule
\end{tabular}

\vspace{1em}

\noindent The following examples are generated on CR.

\vspace{1em}

\begin{tabular}{c|l}
    \toprule
    \textbf{Example 1} & if you are using linux , and you want to use gprs to connect to the net , i suggest this \\
    & phone along with a belkin usb bluetooth adapter . (\textit{positive}) \\
    \textbf{Example 2} & the eax sound enhancement feature did n 't do enough for me and the custom eq was \\
    & a limited 5 band eq . (\textit{negative}) \\
    \midrule
    \textbf{GPT3Mix} & My E550i had 4 months service from Singapore. During those fourmonth I \\
    & recognized that this phone is really lovely, not so popular in Indonesia and nobody use \\
    & like me :-( Anyway today the Phone set become hot when i try to read message... \\
    & ({\textit{positive}}: 22\%, \textbf{\textit{negative}}: 78\%) \\
    \midrule \midrule
    
    \textbf{Example 1} &
        8 battery life is excellent .
    (\textit{positive}) \\
    
    \textbf{Example 2} &
        navigation is a bit tricky and takes getting used to but the included pdf instructions \\ & cover the operation pretty well .
    (\textit{negative}) \\
    
    \midrule
    \textbf{GPT3Mix} & 
        Fast and good GPS ! my favorite is the speed camera app sensitivity > ì. < , you \\ & can threshold values ++to avoid spending penalties on shortcuts which detect it many, \\ & i recommend !!! Good product at a great price!
    (\textbf{\textit{positive}}: 99\%, {\textit{negative}}: 1\%) \\
    
    \bottomrule
\end{tabular}

\vspace{1em}

\noindent The following examples are generated on RT20. Note that GPT-3 would not have learned about some of the movies mentioned in the examples, such as \textit{Tenet}, hence GPT3Mix cannot retrieve knowledge and facts about the movies. This implies that the data augmentation effect is largely derived from linguistic knowledge and diverse perturbations exhibited by GPT-3. Also note that the parentheses in the second example was in the original review. GPT3Mix learns to mix surface patterns in the examples as well.

\vspace{1em}
\begin{tabular}{c|l}
    \toprule
    
    \textbf{Example 1} &
        to be constantly told how tenet's engine works is like having peter jackson explain \\ & the metallurgy behind the one ring -- unnecessary .
    (\textit{negative}) \\
    
    \textbf{Example 2} &
        an alarmist rise-of-the-machines premise points to a `` terminator '' -like future , with \\ & very different results , in this serviceable sci-fi war movie .
    (\textit{positive}) \\
    
    \midrule
    \textbf{GPT3Mix} & 
        a village blacksmith and a feared highwayman form an alliance to star in the offbeat \\ & arrangement of swords .
    (\textbf{\textit{positive}}: 60\%, {\textit{negative}}: 40\%) \\
    
    \midrule \midrule
    
    \textbf{Example 1} &
        the characters are paper-thin , the plot is super predictable , and despite its star- \\ & studded cast only robert pattinson seems like he's having any fun .
    (\textit{negative}) \\
    
    \textbf{Example 2} &
        [ there are ] small moments that exude credibility but go hand in hand with others \\ & that don't .
    (\textit{positive}) \\
    
    \midrule
    \textbf{GPT3Mix} & 
        [ it's ] a robot movie that is fun enough to watch but feels superfluous ; there' s just \\ & not much in here we haven't seen before from superior films like i , robota and \\ & plato 's ashes .
    ({\textit{positive}}: 10\%, \textbf{\textit{negative}}: 90\%) \\
    
    \midrule \midrule
    
    \textbf{Example 1} &
        the only thing that makes the film worth watching is a performance by robert \\ & pattinson , who has been the best thing about quite a few films in the last year . \\ &
    (\textit{negative}) \\
    
    \textbf{Example 2} &
        `` land '' has real power even when stands absolutely still for lengthy periods of \\ & screen time .
    (\textit{positive}) \\
    
    \midrule
    \textbf{GPT3Mix} & 
        the rock will please those who are expecting a visually entertaining . \\ &
    (\textbf{\textit{positive}}: 65\%, {\textit{negative}}: 35\%) \\

    \bottomrule
\end{tabular}
\vspace{1em}

\noindent The following examples are generated on SUBJ. Note that GPT3Mix sometimes struggles to identify the task from the context and predicts the labels with low confidence, as the concept of ``subjectivity'' can be vague and debatable.

\vspace{1em}
\begin{tabular}{c|l}
    \toprule
    
    \textbf{Example 1} &
        smith examines the intimate , unguarded moments of folks who live in unusual homes \\ & -- which pop up in nearly every corner of the country .
    (\textit{subjective}) \\
    
    \textbf{Example 2} &
        this is a film version of the play they wrote based on more than 200 interviews they \\ & conducted in laramie .
    (\textit{objective}) \\
    
    \midrule
    \textbf{GPT3Mix} & 
        reporter covers our corrupt customs laws , and it¹s surprising something isn't done \\ & about them .
    (\textbf{\textit{subjective}}: 59\%, {\textit{objective}}: 41\%) \\
    
    \midrule \midrule
    
    \textbf{Example 1} &
        `` the dangerous lives of altar boys '' has flaws , but it also has humor and heart and \\ & very talented young actors
    (\textit{subjective}) \\
    
    \textbf{Example 2} &
        his family decides to go back on a holiday to india for 2 weeks , when tina discovers \\ &  the truth about pooja's e-mails , they decide together that tina will play along with \\ & the charade .
    (\textit{objective}) \\
    
    \midrule
    \textbf{GPT3Mix} & 
        a rich man hires a hitman for his wife . but she finds out and decides to manipulate \\ & the killer with an ever decreasing budget
    (\textbf{\textit{subjective}}: 49\%, {\textit{objective}}: 51\%) \\

    \bottomrule
\end{tabular}

\end{document}